\documentclass[Afour,sageh,times]{sagej}

\usepackage{moreverb,url}

\usepackage[bookmarksopen,bookmarksnumbered,colorlinks,allcolors=black]{hyperref}

\usepackage{graphicx}
\usepackage{caption}
\usepackage{subcaption}
\usepackage{balance}

\usepackage{siunitx}
\usepackage{amsmath}

\usepackage{authblk}

\global\long\def\mA{\mathtt{A}}
\global\long\def\mB{\mathtt{B}}
\global\long\def\mX{\mathtt{X}}
\global\long\def\mZ{\mathtt{Z}}
\global\long\def\Kint{\mathtt{K}}

\global\long\def\bx{\mathbf{x}}
\global\long\def\bP{\mathbf{P}}
\global\long\def\mT{\mathtt{T}}
\global\long\def\bu{\mathbf{u}}

\begin{document}

\runninghead{The Event-Camera Dataset and Simulator}

\title{The Event-Camera Dataset and Simulator: \\
Event-based Data for Pose Estimation, Visual Odometry, and SLAM}

\author{Elias Mueggler\affilnum{1}, Henri Rebecq\affilnum{1}, Guillermo Gallego\affilnum{1}, Tobi Delbruck\affilnum{2} and Davide Scaramuzza\affilnum{1}}

\affiliation{\affilnum{1}Robotics and Perception Group, University of Zurich, Switzerland\\
\affilnum{2}Institute of Neuroinformatics, University of Zurich and ETH Zurich, Switzerland}

\corrauth{
Elias Mueggler, 
Robotics and Perception Group, 
University of Zurich, Switzerland. 
E-mail: mueggler@ifi.uzh.ch\\[1.2ex]
This work was supported by SNSF-ERC Starting Grant, the DARPA FLA Program, by the Google Faculty Research Award, by the Qualcomm Innovation Fellowship, by the National Centre of Competence in Research Robotics (NCCR), by the Swiss National Science Foundation, and by the UZH Forschungskredit (Grant No: K-34142-03-01).
}

%\email{mueggler@ifi.uzh.ch}

\begin{abstract}
New vision sensors, such as the Dynamic and
Active-pixel Vision sensor (DAVIS), incorporate a conventional
global-shutter camera and an event-based sensor in the same pixel array.
These sensors have great potential for high-speed robotics and computer vision because they
allow us to combine the benefits of conventional cameras
with those of event-based sensors: low latency, high temporal
resolution, and very high dynamic range. However, new algorithms
are required to exploit the sensor characteristics and cope
with its unconventional output, which consists of a stream of
asynchronous brightness changes (called ``events'') and synchronous grayscale frames.
For this purpose, we present and release a collection 
of datasets captured with a DAVIS in a variety of synthetic and real environments, which we hope will motivate research 
on new algorithms for high-speed and high-dynamic-range robotics and computer-vision applications.
In addition to global-shutter intensity images and asynchronous events, 
we provide inertial measurements and ground-truth camera poses from a motion-capture system. 
The latter allows comparing the pose accuracy of ego-motion estimation algorithms quantitatively.
All the data are released both as standard text files and binary files (i.e., rosbag). 
This paper provides an overview of the available data and describes a simulator that we release open-source to create synthetic event-camera data.
\end{abstract}

\keywords{Event-based cameras, visual odometry, SLAM, simulation}

\maketitle

\section*{Dataset Website}
All datasets and the simulator can be found on the web:\\
\url{http://rpg.ifi.uzh.ch/davis_data.html}\\
A video containing visualizations of the datasets:\\
\url{https://youtu.be/bVVBTQ7l36I}\\

\section{Introduction}
Over the past fifty years, computer-vision research has been devoted to standard, frame-based cameras 
(i.e., rolling or global shutter cameras) and only in the last few years cameras have been
successfully used in commercial autonomous mobile robots, such as cars, drones, and vacuum cleaners, just to mention a few.
Despite the recent progress, we believe that the advent of event-based cameras is about to revolutionize the robot sensing landscape.
Indeed, the performance of a mobile robot in tasks, such as navigation, depends on the accuracy and latency of perception. 
The latency depends on the frequency of the sensor data plus the time it takes to process the data. 
It is typical in current robot-sensing pipelines to have latencies in the order of \SIrange{50}{200}{\milli\second} or more, which puts a hard bound on 
the maximum agility of the platform. 
An event-based camera virtually eliminates the latency: data is transmitted 
using events, which have a latency in the order of \emph{micro-}seconds. 
Another advantage of event-based cameras is their very high dynamic range (\SI{130}{\decibel} vs. \SI{60}{\decibel} of standard cameras), which makes
them ideal in scenes characterized by large illumination changes.
Other key properties of event-based cameras are low-bandwidth, low-storage, and low-power requirements. 
All these properties enable the design of a new class of algorithms for high-speed and high-dynamic-range robotics, 
where standard cameras are typically not ideal because of motion blur, image saturation, and high latency. 
However, the way that event-based cameras convey the information is completely different from that of traditional sensors, 
so that a paradigm shift is needed to deal with them.

\begin{figure}[t!]
  \centering
  \begin{subfigure}[b]{0.9\linewidth}
  	\includegraphics[width=\linewidth]{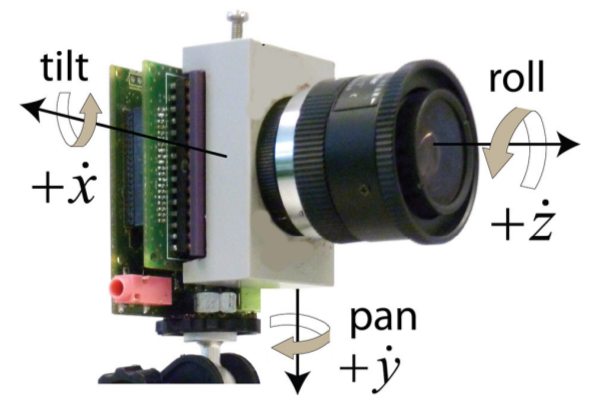}
  	\caption{The DAVIS sensor and axes definitions. Figure adapted from~\citep{Delbruck14iscas}}
  	\label{fig:davis_axes}
  \end{subfigure}
  
  \vspace{1em}
  
  \begin{subfigure}[b]{0.9\linewidth}
    \includegraphics[width=\linewidth]{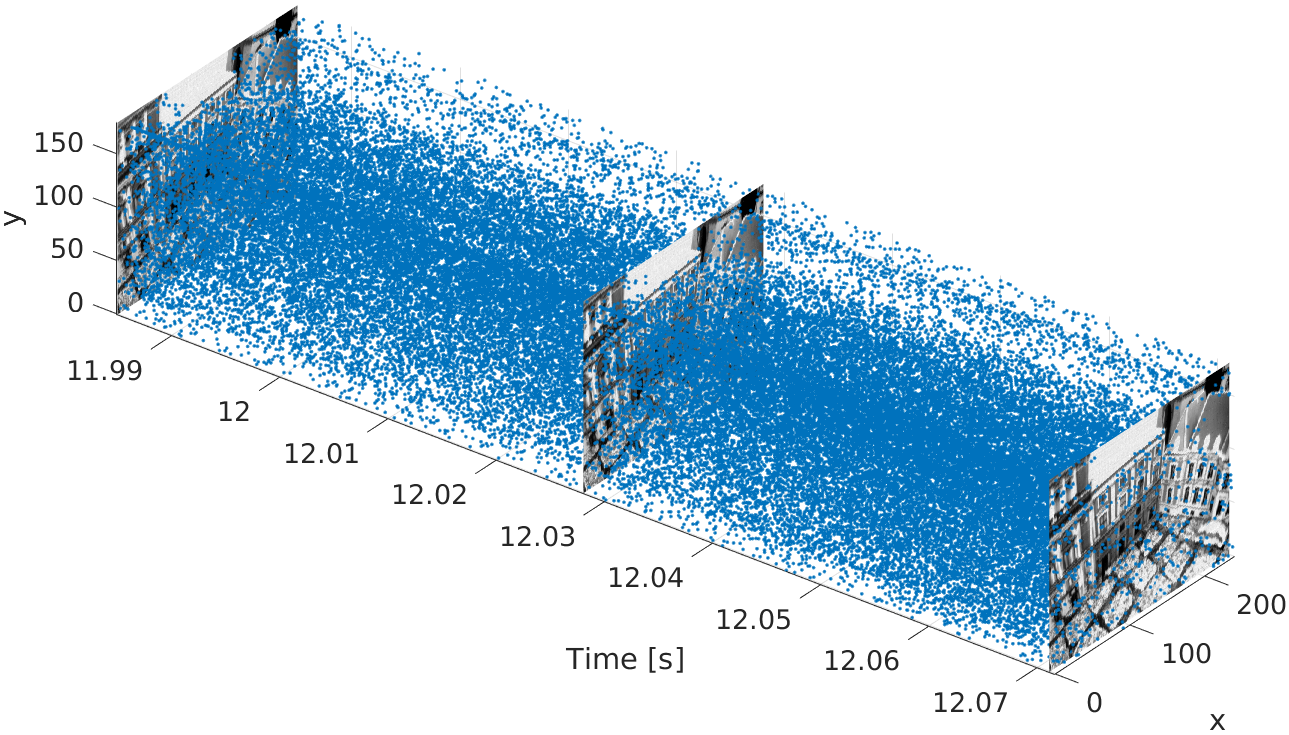}
  \caption{Visualization of the event output of a DAVIS in space-time. 
  		Blue dots mark individual asynchronous events.
		The polarity of the events is not shown.}
    \label{fig:dvs_output}
  \end{subfigure}
  
  \caption{The DAVIS camera and visualization of its output.}
  \label{fig:davis_output}
\end{figure}

\subsection{Related Datasets}
There exist two recent datasets that also use the DAVIS: \citep{Rueckauer16fns} and \citep{Barranco16fns}.

The first work is tailored for comparison of event-based optical flow estimation algorithms~\citep{Rueckauer16fns}.
It contains both synthetic and real datasets under pure rotational (3 degrees of freedom (DOF)) motion on simple scenes with strong visual contrasts. 
Ground truth was acquired using the inertial measurement unit (IMU).
In contrast, our datasets contain arbitrary, hand-held, 6-DOF motion in a variety of artificial and natural scenes with precise ground-truth camera poses from a motion-capture system.

A more similar work to ours is~\citep{Barranco16fns}. 
Their focus is to create a dataset that facilitates comparison of event-based and frame-based methods for 2D and 3D visual navigation tasks.
To this end, a ground robot was equipped with a DAVIS and a Microsoft Kinect \mbox{RGB-D} sensor.
The DAVIS was mounted on a pan-tilt unit, thus it could be excited in 5-DOF.
The scene contains checkerboards, books, and a chair.
Ground truth was acquired by the encoders of pan-tilt unit and the ground robot's wheel odometry, and is therefore subject to drift. 
In contrast, our dataset contains hand-held, 6-DOF motion (slow- and high-speed) on a variety of scenes with precise ground-truth camera poses from a motion-capture system, which is not subject to drift.

\section{The DAVIS Sensor}

The Dynamic and Active-pixel Vision Sensor (DAVIS)~\citep{Brandli2014ssc} (see Fig.~\ref{fig:davis_axes}) is an event camera that transmits \emph{events} in addition to frames. 
Events are pixel-level, relative-brightness changes that are detected in continuous time by specially-designed pixels\footnote{Video illustration: \url{https://youtu.be/LauQ6LWTkxM}}.
The events are timestamped with \emph{micro-}second resolution and transmitted asynchronously at the time they occur.
Each event $e$ is a tuple $\langle x, y, t, p \rangle$, where $x, y$ are the pixel coordinates of the event, $t$ is the timestamp of the event, and $p=\pm 1$ is the polarity of the event, which is the sign of the brightness change. 
This representation is sometimes also referred to as Address-Event Representation (AER).
The DAVIS has a spatial resolution of $240 \times 180$ pixels.
A visualization of the event output is shown in Fig.~\ref{fig:dvs_output}.
Both the events and frames are generated by the same physical pixels, hence there is no spatial offset between the events and the frames. 

Due to its low latency and high temporal resolution, both in the range of \textit{micro}-seconds, event-based cameras are very promising sensors for high-speed mobile robot applications.
Since event cameras are data-driven (only brightness \textit{changes} are transmitted), no redundant data is transmitted.
The required bandwidth thus depends on the motion speed and the type of scene.
An additional advantage for robotic applications is the high dynamic range of~\SI{130}{\decibel} (compared to~\SI{60}{\decibel} of expensive computer-vision cameras), which allows both indoor and outdoor operation without changing parameters.
Since all pixels are independent, very large contrast changes can also take place within the same scene.

Over the course of the last seven years, several groups including ours have demonstrated the use of event-based sensors in a variety of tasks, such as SLAM in 2D~\citep{Weikersdorfer13icvs} and 3D~\citep{Kueng16iros,Kim16eccv,Rebecq16ral}, 
optical flow~\citep{Cook11ijcnn,Benosman14tnnls,Bardow16cvpr}, 
visual odometry~\citep{Censi14icra}, 
\mbox{6-DOF} localization for high-speed robotics~\citep{Mueggler14iros}, 
line detection and localization~\citep{Yuan16icra},
3D reconstruction~\citep{Rebecq16bmvc},
image reconstruction and mosaicing~\citep{Kim14bmvc,Reinbacher16bmvc}, 
orientation estimation~\citep{Gallego16ral}, and
continuous-time trajectory estimation~\citep{Mueggler15rss}.

However, all these methods were evaluated on different, specific datasets and, therefore, cannot be compared against each other. 
The datasets we propose here are tailored to allow comparison of pose tracking, visual odometry, and SLAM algorithms. 
Since event-based cameras, such as the DAVIS, are currently still expensive ($\sim 5,000$ USD), these data also allow researchers without equipment to use well-calibrated data for their research.

\subsection{DAVIS IMU} 
In addition to the visual output (events and frames), 
the DAVIS includes an IMU that provides gyroscope and accelerometer data, 
thus enabling to design visual-inertial event-based algorithms.
The DAVIS camera has the IMU mounted directly behind and centered under the image sensor pixel array center, 
at a distance of about \SI{3}{\milli\meter} from it, 
so that the IMU shares nearly the same position as the event sensor (i.e., the photoreceptor, not the optical center of the camera, since this is lens dependent; 
the camera-IMU calibration is discussed on page~\pageref{sec:calib-camera-imu}). 
The IMU axes are aligned with the visual sensor axes (see Fig.~\ref{fig:davis_axes}). 
More specifically, the IMU is an InvenSense MPU-6150\footnote{IMU data sheet: https://store.invensense.com/ProductDetail/MPU6150-invensense/470090/}, 
which integrates a three-axis gyroscope that can measure in the range $\pm \SI{2000}{\degree/\second}$
and a three-axis accelerometer for the range $\pm 16 g$.
It integrates six 16-bit ADCs for digitizing the gyroscope and accelerometer outputs at \SI{1}{\kilo\hertz} sample rate.

\section{DAVIS Simulator}
\begin{figure}[t!]
  \centering
  \includegraphics[width=\columnwidth]{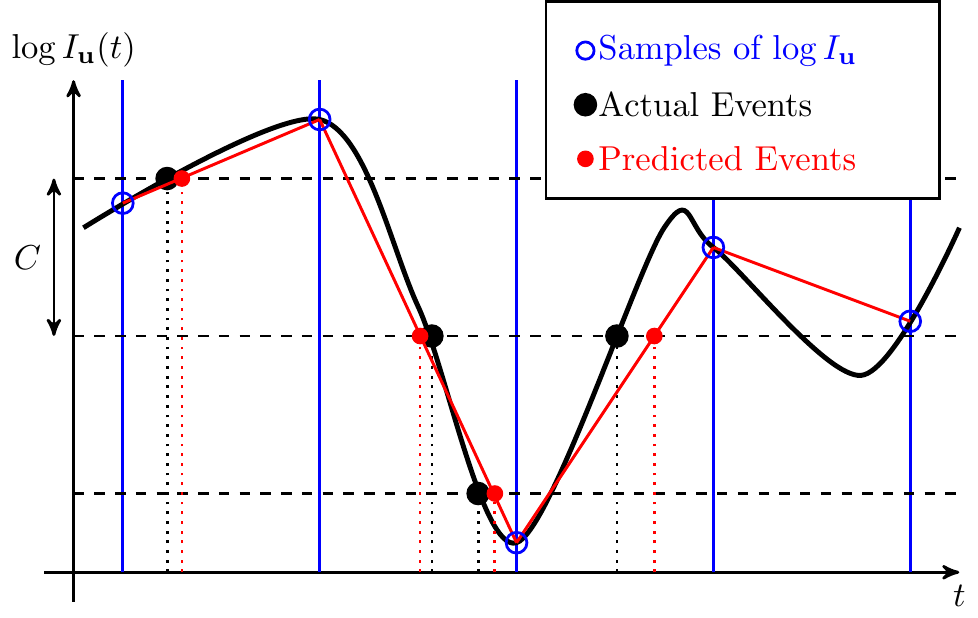}
  \caption{DAVIS Simulator. 
  Per-pixel event generation using piecewise linear time interpolation of the intensities given by the rendered images. 
  For simplicity, images were rendered at a fixed rate.
  }
  \label{fig:simulator}
\end{figure}

Simulation offers a good baseline when working with new sensors, such as the DAVIS.
Based on the operation principle of an ideal DAVIS pixel, we created a simulator that, given a virtual 3D scene and the trajectory of a moving DAVIS within it, generates the corresponding stream of events, intensity frames, and depth maps.
We used the computer graphics software Blender\footnote{https://www.blender.org/} to generate thousands of rendered images along the specified trajectory, ensuring that the motion between consecutive images was smaller than $1/3$ pixel.
For each pixel, we keep track of the time of the last event triggered at that location. 
This map of timestamps (also called surface of active events~\citep{Benosman14tnnls}), combined with time interpolation of the rendered image intensities, allows determining brightness changes of predefined amount (given by the contrast threshold) in the time between images, thus effectively providing continuous timestamps, as if events were generated asynchronously.
Time interpolation has an additional benefit: it solves the problem of having to generate millions of images for each second of a sequence, as it would have been required to deliver microsecond-resolution timestamps in the absence of interpolation.

More specifically, Fig.~\ref{fig:simulator} illustrates the operation of the simulator for a single pixel $\bu=(x,y)^\top$.
The continuous intensity signal at pixel $\bu$, $\log I_\bu(t)$ (black) is sampled at the times of the rendered images (blue markers).
These samples are used to determine the times of the events: 
the data is linearly interpolated between consecutive samples 
and the crossings of the resulting lines (in red) with the levels given by multiples of the contrast threshold $C$ (i.e., horizontal lines) 
specify the timestamps of the events (red dots).
As it can be observed, this simple interpolation scheme allows for
$(i)$ higher resolution event time stamps than those of the rendered images, 
and $(ii)$ the generation of multiple events between two samples if the corresponding intensity jump is larger than the contrast threshold.

The provided events are ``perfect'' measurements up to sampling and quantization;
under this condition, an image $\hat{I}(\bu;t)$ can be reconstructed from the event stream at any point in time $t$ by accumulating events $e_k = \langle \bu_k, t_k, p_k \rangle$ according to
$$\log \hat{I}(\bu;t) = \log I(\bu;0) + \sum_{0 < t_k\leq t} p_k \,C\, \delta(\bu-\bu_k)\delta(t-t_k),$$ 
where $I(\bu;0)$ is the rendered image at time $t=0$ and $\delta$ selects the pixel to be updated on every event (pixel $\bu_k$ of $\hat{I}$ is updated at time $t_k$).
We used this scheme to check that the reconstructed image agreed with the rendered image at several points in time; specifically, the per-pixel intensity error was confined to the quantization interval $(-C,C)$.

Event generation operates on brightness pixels, which are computed from the rendered color images 
using the ITU-R Recommendation BT.601\footnote{https://www.itu.int/rec/R-REC-BT.601} for luma, i.e., 
according to formula $Y = 0.299 R + 0.587 G + 0.114 B$, with RGB channels in linear color space to better resemble the operation of the DAVIS.

Because realistic event noise is extremely difficult to model due to the complex behavior of event sensors with respect to their bias settings and other factors, the provided simulation datasets do not include event noise. 
Nevertheless, the simulator, and the datasets created with it, are a useful tool for prototyping new event-based algorithms.
Our implementation is available as open-source software.%
\footnote{https://github.com/uzh-rpg/rpg\_davis\_simulator}

\section{Datasets}

\begin{figure*}[ht!]
    \centering
    \begin{subfigure}[b]{0.31\textwidth}
        \includegraphics[width=\textwidth]{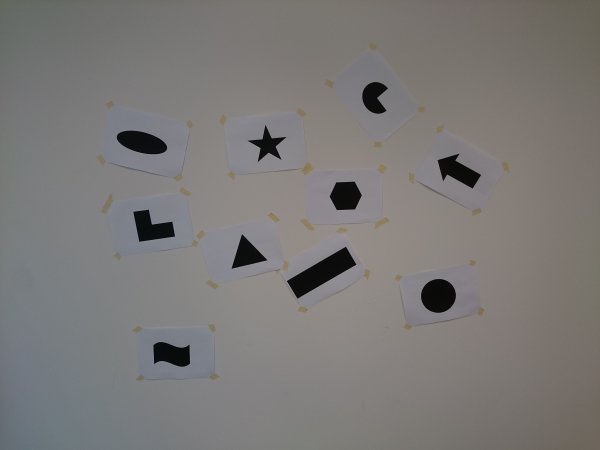}
        \caption{Shapes}
        \label{fig:scenes:shapes}
    \end{subfigure}
    ~
    \begin{subfigure}[b]{0.31\textwidth}
        \includegraphics[width=\textwidth]{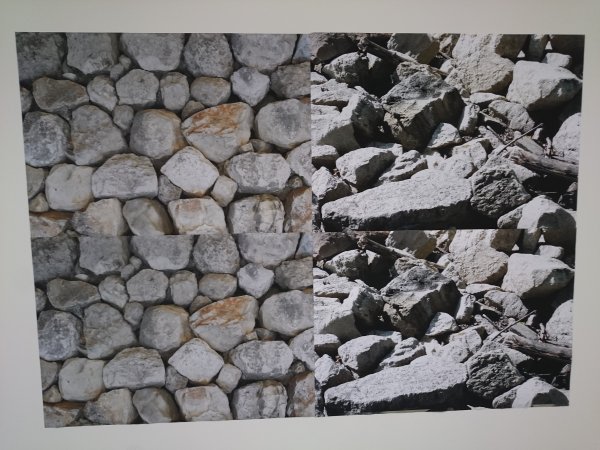}
        \caption{Wall Poster}
        \label{fig:scenes:poster}
    \end{subfigure}
    ~ 
    \begin{subfigure}[b]{0.31\textwidth}
        \includegraphics[width=\textwidth]{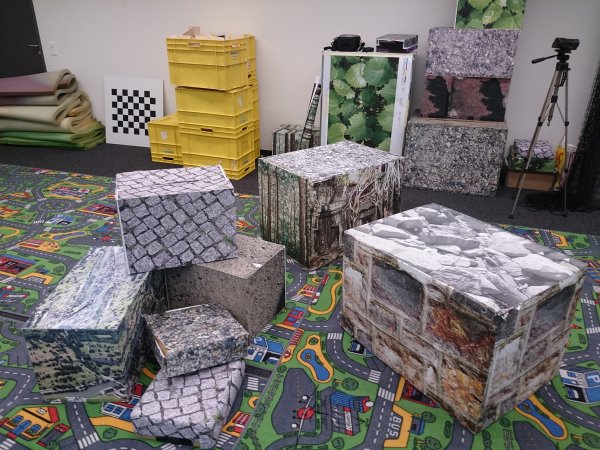}
        \caption{Boxes}
        \label{fig:scenes:boxes}
    \end{subfigure}
    
    \vspace{0.5em}
    
    \begin{subfigure}[b]{0.31\textwidth}
        \includegraphics[width=\textwidth]{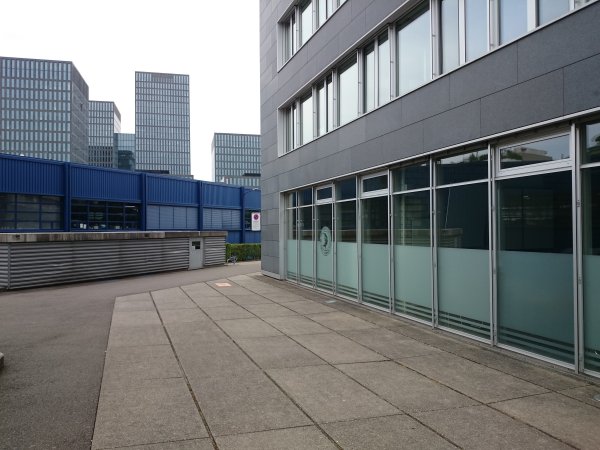}
        \caption{Outdoors}
        \label{fig:scenes:outdoors}
    \end{subfigure}
    ~
    \begin{subfigure}[b]{0.31\textwidth}
        \includegraphics[width=\textwidth]{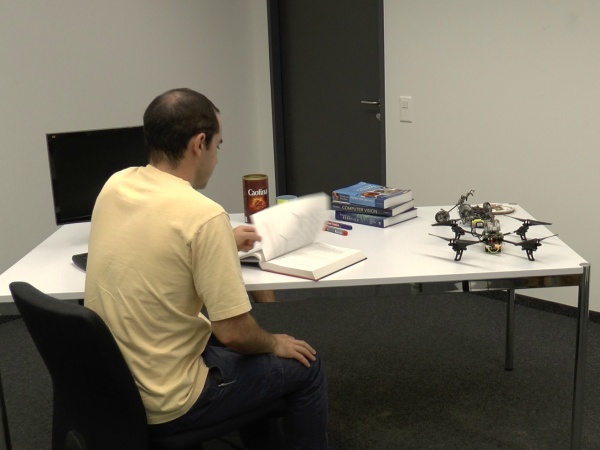}
        \caption{Dynamic}
        \label{fig:scenes:dynamic}
    \end{subfigure}
    ~
    \begin{subfigure}[b]{0.31\textwidth}
        \includegraphics[width=\textwidth]{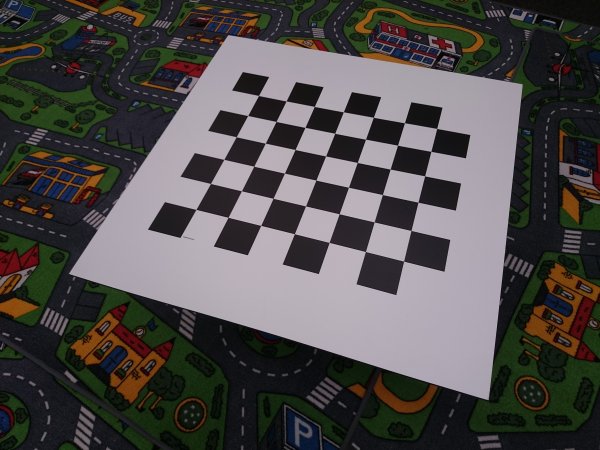}
        \caption{Calibration}
        \label{fig:scenes:calibration}
    \end{subfigure}
    
    \vspace{0.5em}
    
    \begin{subfigure}[b]{0.31\textwidth}
        \includegraphics[width=\textwidth]{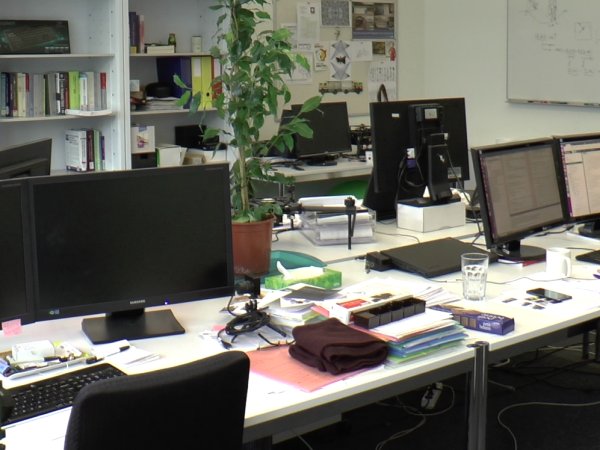}
        \caption{Office}
        \label{fig:scenes:office}
    \end{subfigure}
    ~    
    \begin{subfigure}[b]{0.31\textwidth}
        \includegraphics[width=\textwidth]{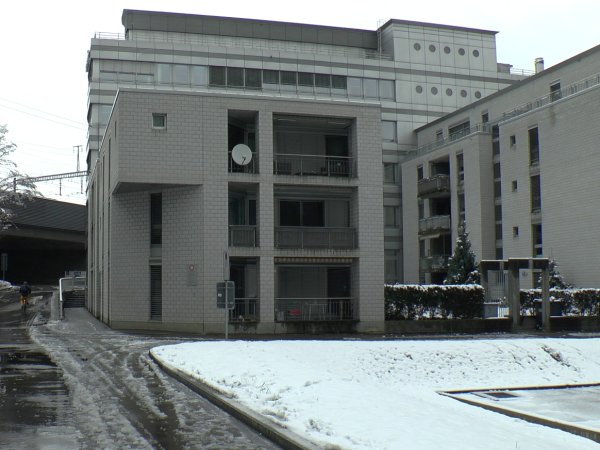}
        \caption{Urban}
        \label{fig:scenes:urban}
    \end{subfigure}
    ~
    \begin{subfigure}[b]{0.31\textwidth}
        \includegraphics[width=\textwidth]{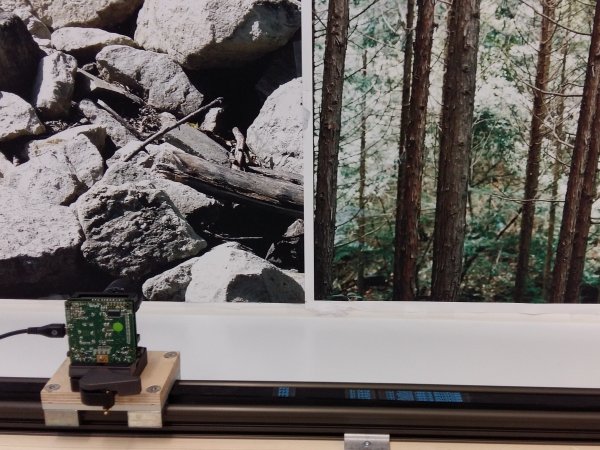}
        \caption{Motorized linear slider}
        \label{fig:scenes:slider}
    \end{subfigure}
    
    \vspace{0.5em}    
    
    \begin{subfigure}[b]{0.31\textwidth}
        \includegraphics[width=\textwidth]{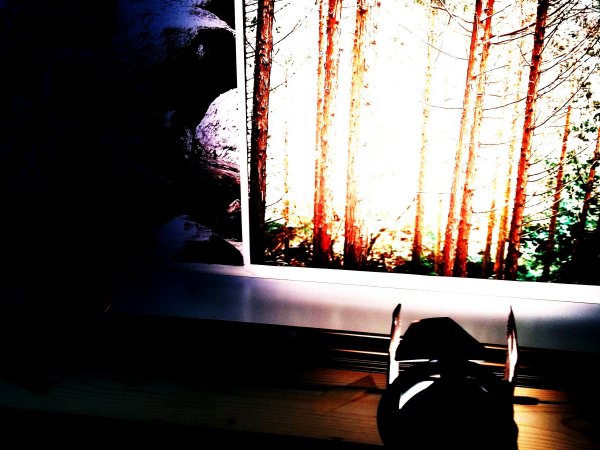}
        \caption{Motorized slider (HDR)}
        \label{fig:scenes:slider_hdr}
    \end{subfigure}
    ~
    \begin{subfigure}[b]{0.31\textwidth}
        \includegraphics[width=\textwidth]{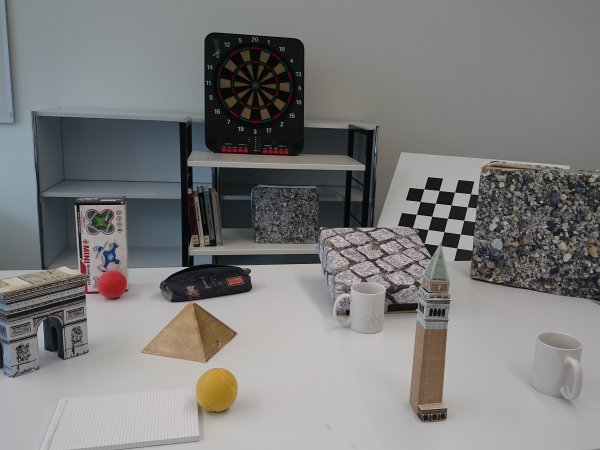}
        \caption{Motorized slider with objects}
        \label{fig:scenes:slider_depth}
    \end{subfigure}
    ~
    \begin{subfigure}[b]{0.31\textwidth}
        \includegraphics[width=\textwidth]{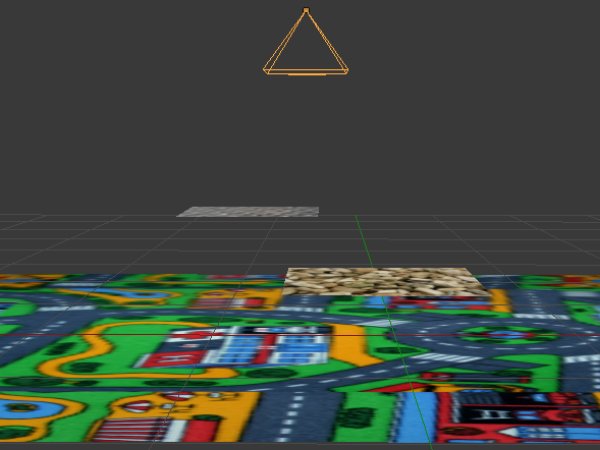}
        \caption{Synthetic: 3 planes}
        \label{fig:scenes:simulation_3planes}
    \end{subfigure}
    
    \vspace{0.5em}
    
    \begin{subfigure}[b]{0.31\textwidth}
        \includegraphics[width=\textwidth]{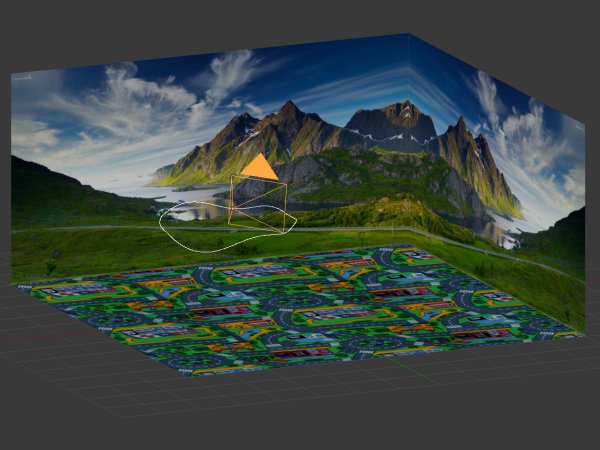}
        \caption{Synthetic: 3 walls}
        \label{fig:scenes:simulation_3walls}
    \end{subfigure}
    
    \caption{Dataset scenes}\label{fig:scenes}
\end{figure*}

In this section, we describe the datasets that we provide.
The datasets contain: 
\begin{itemize}
  \item the asynchronous event stream,
  \item intensity images at about \SI{24}{\hertz},
  \item inertial measurements (3-axis gyroscope and 3-axis accelerometer) at \SI{1}{\kilo\hertz},
  \item ground-truth camera poses from a motion-capture system\footnote{We use an OptiTrack system from NaturalPoint.} with sub-millimeter precision at \SI{200}{\hertz} (for the indoor datasets),
  \item the intrinsic camera matrix.
\end{itemize}
All information comes with precise timestamps.
For datasets that were captured outside the motion-capture system (e.g., in an office or outdoors), no ground truth is provided.
Some datasets were collected using a motorized linear slider and ground truth was collected using the slider's position.
Due to vibrations induced by the slider motor, the very noisy IMU data was not recorded.

\subsection{Data Format}

\begin{table*}[t!]
\centering 
\begin{tabular}{| l | l | l |}
\hline
\textbf{File} & \textbf{Description} & \textbf{Line Content} \\
\hline
\texttt{events.txt} & One event per line & \texttt{timestamp x y polarity} \\
\texttt{images.txt} & One image reference per line & \texttt{timestamp filename} \\
\texttt{images/00000000.png} &  Images referenced from \texttt{images.txt} & \\
\texttt{imu.txt} & One measurement per line  & \texttt{timestamp ax ay az gx gy gz} \\
\texttt{groundtruth.txt} & One ground-truth measurement per line  & \texttt{timestamp px py pz qx qy qz qw} \\
\texttt{calib.txt} & Camera parameters & \texttt{fx fy cx cy k1 k2 p1 p2 k3} \\
\hline
\end{tabular}
\caption{Description of Dataset Format}
\label{table:format}
\end{table*}

The datasets are provided in standard text form that is described here.
For convenience, they can also be downloaded as binary rosbag files (the details are on the website).
The format of the text files is described in Table~\ref{table:format}.

The ground-truth pose is with respect to the (arbitrary) motion-capture origin that has the $z$-axis gravity-aligned (pointing upwards).
The orientation is provided as a unit quaternion $\mathbf{q} = (q_x,q_y,q_z,q_w)^\top$, 
where $q_w$ and $\mathbf{q}_v = (q_x,q_y,q_z)^\top$ are the scalar and vector components, respectively.
This convention was proposed as a standard by JPL~\citep{Breckenridge79jpl}.

All values are reported in SI units.
While the timestamps were originally recorded as POSIX, we subtracted the lowest timestamp as offset such that all datasets start at zero.
This helps to avoid numerical difficulties when dealing with microsecond resolution timestamps of the events.

Images are provided as PNG files.
The list of all images and their timestamps is provided in a separate file.
The typical framerate is \SI{24}{\hertz}, but it varies with the exposure time.

The IMU axes have the same orientation as those of the optical coordinate frame (i.e., the positive $z$-axis is aligned with the optical axis and so are the $x$- and $y$-axes).

\subsection{List of Datasets}
The provided datasets are summarized in Table~\ref{table:datasets} and Fig.~\ref{fig:scenes}.
All the datasets contain increasing speeds, different scenes, and varying degrees of freedom\footnote{The DAVIS was moved by hand, the dominant motion is described.}:
for the \texttt{shapes}, \texttt{poster}, and \texttt{boxes} datasets, the motion first starts with excitation of each single degree of freedom separately;
then combined and faster excitations are performed. 
This leads to increasing difficulty and a higher event rate over time.

\begin{table*}[t!]
\centering 
\begin{tabular}{| l | l | l | l | S[table-format=3.1, round-mode=places, round-precision=1] | S[table-format=1.2, round-mode=places, round-precision=2] | S[table-format=3.0, round-mode=places, round-precision=0] | S[table-format=9.0] |}
\hline
\textbf{Name} & \textbf{Motion} & \textbf{Scene} & \textbf{GT} & \textbf{T [\si{\second}]} & \textbf{TS [\si{\meter/\second}]} & \textbf{RS [\si{\degree/\second}]} & \textbf{NE [-]} \\
\hline
\texttt{shapes\_rotation} & Rotation, incr. speed & Fig.~\ref{fig:scenes:shapes} & yes &             59.787844 & 0.83 & 730.17 & 23126288 \\
\texttt{shapes\_translation} & Translation, incr. speed & Fig.~\ref{fig:scenes:shapes} & yes &       59.734202 & 2.6 & 270.78 & 17363976 \\
\texttt{shapes\_6dof} & 6 DOF, incr. speed & Fig.~\ref{fig:scenes:shapes} & yes &                    59.72497 & 2.35 & 714.78 & 17962477 \\
\hline
\texttt{poster\_rotation} & Rotation, incr. speed & Fig.~\ref{fig:scenes:poster} & yes &             59.791966 & 0.84 & 884.15 & 169350136 \\
\texttt{poster\_translation} & Translation, incr. speed & Fig.~\ref{fig:scenes:poster} & yes &       59.790562 & 2.58 & 240.46 & 100033286 \\
\texttt{poster\_6dof} & 6 DOF, incr. speed & Fig.~\ref{fig:scenes:poster} & yes &                    59.832581 & 2.51 & 937.22 & 133464530 \\
\hline
\texttt{boxes\_rotation} & Rotation, incr. speed & Fig.~\ref{fig:scenes:boxes} & yes &               59.818996 & 0.85 & 669.12 & 185688947 \\
\texttt{boxes\_translation} & Translation, incr. speed & Fig.~\ref{fig:scenes:boxes} & yes &         59.776029 & 3.43 & 318.59 & 112388307 \\
\texttt{boxes\_6dof} & 6 DOF, incr. speed & Fig.~\ref{fig:scenes:boxes} & yes &                      59.772908 & 3.84 & 508.51 & 133085511 \\
\hline
\texttt{hdr\_poster} & 6 DOF, incr. speed & Fig.~\ref{fig:scenes:poster} & yes &              59.77225 & 2.28 & 597.09 & 102910720 \\
\texttt{hdr\_boxes} & 6 DOF, incr. speed & Fig.~\ref{fig:scenes:boxes} & yes &                59.780373 & 2.94 & 591.71 & 118499744 \\
\hline
\texttt{outdoors\_walking} & 6 DOF, walking & Fig.~\ref{fig:scenes:outdoors} & no$^\dagger$ &       133.434313 & n/a & n/a & 64517638 \\
\texttt{outdoors\_running} & 6 DOF, running & Fig.~\ref{fig:scenes:outdoors} & no$^\dagger$ &        87.619327 & n/a & n/a & 98572164 \\
\hline
\texttt{dynamic\_rotation} & Rotation, incr. speed & Fig.~\ref{fig:scenes:dynamic} & yes &             59.759053 & 0.45 & 542.29 &  71324510 \\
\texttt{dynamic\_translation} & Translation, incr. speed & Fig.~\ref{fig:scenes:dynamic} & yes &       59.798989 & 1.86 & 227.2  &  35809924 \\
\texttt{dynamic\_6dof} & 6 DOF, incr. speed & Fig.~\ref{fig:scenes:dynamic} & yes &                    59.730322 & 2.91 & 627.12 &  57174637 \\
\hline
\texttt{calibration} & 6 DOF, slow & Fig.~\ref{fig:scenes:calibration} & yes &                       59.791377 & 0.32 & 67.21 & 21340629 \\
\hline
\hline
\texttt{office\_zigzag} & 6-DOF, zigzag, slow & Fig.~\ref{fig:scenes:office} & no &                 10.914726 & n/a & n/a & 7735308 \\
\texttt{office\_spiral} & 6-DOF, spiral, slow & Fig.~\ref{fig:scenes:office} & no &                 11.180662 & n/a & n/a & 6254774 \\
\hline
\texttt{urban} & Linear, slow & Fig.~\ref{fig:scenes:urban} & no &                                   10.730572 & n/a & n/a & 5359539 \\
\hline
\texttt{slider\_close} & Linear, const, speed & Fig.~\ref{fig:scenes:slider} & yes\textsuperscript{*} &                 6.459701 & 0.16 & 0 & 4032668 \\
\texttt{slider\_far} & Linear, const, speed & Fig.~\ref{fig:scenes:slider} & yes\textsuperscript{*} &                 6.396429 & 0.16 & 0 & 3442683 \\
\texttt{slider\_hdr\_close} & Linear, const. speed & Fig.~\ref{fig:scenes:slider_hdr} & yes\textsuperscript{*} &             6.53924 & 0.16 & 0 & 3337787 \\
\texttt{slider\_hdr\_far} & Linear, const. speed & Fig.~\ref{fig:scenes:slider_hdr} & yes\textsuperscript{*} &             6.467819 & 0.16 & 0 & 2509582 \\
\texttt{slider\_depth} & Linear, const. speed & Fig.~\ref{fig:scenes:slider_depth} & yes\textsuperscript{*}  & 3.386478 & 0.32 & 0 & 1078541 \\
\hline
\texttt{simulation\_3planes} & Translation, circle & Fig.~\ref{fig:scenes:simulation_3planes} & yes\textsuperscript{\#}      &  2.0 & 0.63 & 0 & 6870278 \\
\texttt{simulation\_3walls} & 6 DOF & Fig.~\ref{fig:scenes:simulation_3walls} & yes\textsuperscript{\#}  & 2.0 & 5.31 & 109.02 & 4104833 \\
\hline
\end{tabular}
\caption{List of Datasets. Note that the calibration dataset only applies to the upper half of the table. The other datasets use different lenses and calibrations.
  GT: Ground truth. T: Duration. TS: Maximum translation speed. RS: Maximum rotational speed. NE: Number of events.
  $^\dagger$Same start and end pose after a large loop.
  \textsuperscript{*}Ground truth from motorized linear slider. No IMU data due to vibrations.
  \textsuperscript{\#}Simulated DAVIS using Blender. No IMU data included.
  }
\label{table:datasets}
\end{table*}

In the high-dynamic-range (HDR) sequences (\texttt{hdr\_poster}, \texttt{hdr\_boxes}, and \texttt{slider\_hdr}), a spotlight was used to create large intrascene contrasts. 
For \texttt{hdr\_poster}, we measured \SI{80}{\lux} and \SI{2400}{\lux} in the dark and bright areas, respectively.

The \texttt{outdoors} datasets were acquired in an urban environment both walking and running.
While no ground truth is available, we returned precisely to the same location after a large loop.

The \texttt{dynamic} datasets were collected in a mock-up office environment viewed by the motion-capture system, with a moving person first sitting at a desk, then moving around.

A \texttt{calibration} dataset is also available, for instance in case the user wishes to use a different camera model or different methods for hand-eye calibration.
The dimensions of the calibration pattern (a checkerboard) are $6 \times 7$ tiles of \SI{70}{\milli\meter}.
For the lower half of the table, different settings (lenses, focus, etc.) were used.
Thus, while we provide the intrinsic calibration, no calibration datasets are available.

The \texttt{slider\_close}, \texttt{slider\_far}, \texttt{slider\_hdr\_close}, and \texttt{slider\_hdr\_far} datasets were recorded with a motorized linear slider parallel to a textured wall at \SI{23.1}{\centi\meter}, \SI{58.4}{\centi\meter}, \SI{23.2}{\centi\meter}, and \SI{58.4}{\centi\meter}, respectively.

For the datasets, we applied two different sets of biases (parameters) for the DAVIS, as listed in Table~\ref{table:biases}.
The first set, labeled ``indoors'', was used in all datasets but \texttt{outdoors\_walking}, \texttt{outdoors\_running}, and \texttt{urban}, where the set ``outdoors'' was applied.
For the simulated datasets, we used a contrast threshold of \SI{\pm 15}{\percent} and \SI{\pm 20}{\percent} for the \texttt{simulation\_3planes} and \texttt{simulation\_3walls}, respectively.

For the simulated scenes, we also provide the 3D world model in Blender (cf. Fig.~\ref{fig:scenes:simulation_3planes} and~\ref{fig:scenes:simulation_3walls}).
In addition to the intensity images and events, these datasets include a depth map for each image frame at~\SI{40}{\hertz}, encoded as 32-bit floating-point values (in the OpenEXR data format).

\begin{table}[h!]
\centering 
\begin{tabular}{| l | S | S | S | S |}
\hline
\textbf{Bias} & \multicolumn{2}{c|}{\textbf{Indoors}} & \multicolumn{2}{c|}{\textbf{Outdoors}} \\
 & {Coarse} & {Fine} & {Coarse} & {Fine} \\
\hline
\texttt{DiffBn} & 2 &  39 & 4 &  39 \\
\texttt{OFFBn}  & 1 &  62 & 4 &   0 \\
\texttt{ONBn}   & 4 & 200 & 6 & 200 \\
\texttt{PrBp}   & 3 &  72 & 2 &  58 \\
\texttt{PrSFBp} & 3 &  96 & 1 &  33 \\
\texttt{RefrBp} & 3 &  52 & 4 &  25 \\
\hline
\end{tabular}
\caption{List of biases applied to the DAVIS. The DAVIS uses two stages of biases, coarse and fine, which we report here.}
\label{table:biases}
\end{table}

\begin{figure}
  \centering
    \includegraphics[width=\columnwidth]{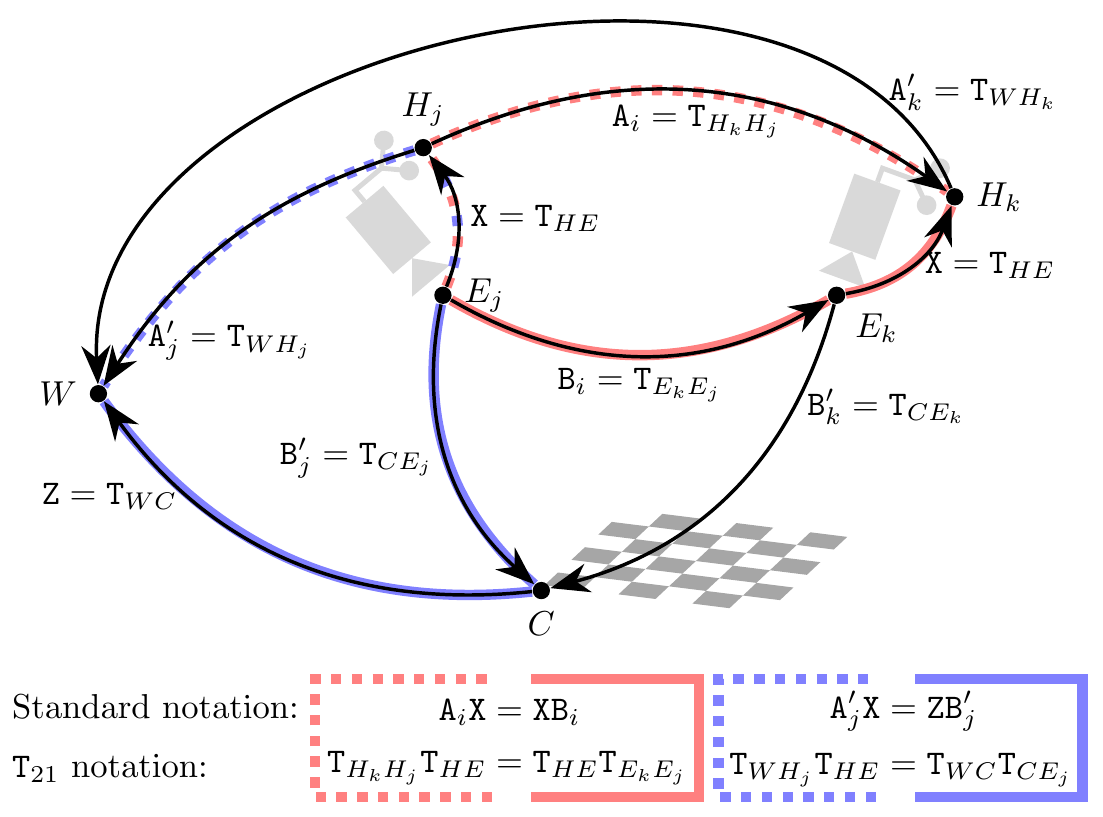}
  \caption{Hand-eye calibration. 
  Coordinate frames and transformations involved in case of the hand-eye device at two different positions ($j$ and $k$).
  The red loop between two stations of the hand-eye device is used in the first type of hand-eye calibration problems, of the form $\mA_i\mX=\mX\mB_i$, and the blue loop is used in the second type of hand-eye calibration problems, of the form $\mA'_j\mX=\mZ\mB'_j$. 
  We use a combination of both approaches to solve for the constant, hand-eye calibration transform $\mX$.
  }
  \label{fig:hand_eye}
\end{figure}

\section{Calibration}
First, we calibrated the DAVIS intrinsically using a checkerboard pattern.
Then, we computed the hand-eye calibration that we applied to the subsequent dataset recordings so that the ground-truth poses that we provide are those of the event camera (i.e., the ``eye''), not those of the motion-capture trackable (i.e., the ``hand'') attached to the camera. 
We also included a calibration dataset in case a different camera model or improved hand-eye calibration method is required.

\subsection{Intrinsic Camera Calibration}
We used the standard pinhole camera model with radial-tangential distortion using the implementation of ROS and OpenCV\footnote{\url{http://wiki.ros.org/camera_calibration/Tutorials/MonocularCalibration}}.
We used three radial distortion coefficients ($k_1$, $k_2$, and $k_3=0$) and two for tangential distortion ($p_1$ and $p_2$).
The distortion coefficients are listed in \texttt{calib.txt} in the same order as in OpenCV.
We provide a dataset for post-calibration in case that another method is preferred.

\subsection{Hand-Eye Calibration}
For the indoor datasets, we provide accurate and high-frequency (\SI{200}{\hertz}) pose data from a motion-capture system.
However, the coordinate frame used by the motion-capture system is different from the optical coordinate frame of the DAVIS.
Thus, we performed a hand-eye calibration before acquiring the datasets.
Fig.~\ref{fig:hand_eye} shows the coordinate frames and transformations used to solve the hand-eye calibration problem.
The frames are those of the world $W$, the hand $H$, the camera $E$ (Fig.~\ref{fig:davis_axes}), and the checkerboard $C$.
For the transformations, Fig.~\ref{fig:hand_eye} shows both the compact standard notation of hand-eye calibration problems and a more explicit one:
the Euclidean transformation $\mT_{ba}$ ($4\times4$ homogeneous matrix representation) maps points from frame $a$ to frame $b$ according to $\bP_{b} = \mT_{ba}\bP_{a}$.

More specifically, we first use a linear algorithm~\citep{Tsai89tro} to provide an initial solution of the hand-eye calibration problem $\{\mA_i\mX=\mX\mB_i\}_{i=1}^N$,
where ${\mA_i\leftrightarrow\mB_i}$ are 
$N$ correspondences of relative hand-hand ($\mA_i:=\mT_{H_k H_j}$) and eye-eye ($\mB_i:=\mT_{E_k E_j}$) poses at different times ($j$ and $k$), respectively, 
and $\mX:=\mT_{HE}$ is the unknown eye-to-hand transformation.
Then, using the second formulation of hand-eye calibration problems, of the form $\{\mA'_j\mX=\mZ\mB'_j\}_{j=1}^{N+1}$,
where $\mA'_j := \mT_{W H_j}$ and $\mB'_j := \mT_{C E_j}$ are the hand-to-motion-capture and eye-to-checkerboard transformations for the \mbox{$j$-th} pose, respectively,
we refined $\mT_{HE}$ by jointly estimating the hand-eye $\mX$ and robot-world $\mZ:=\mT_{WC}$ (i.e., motion-capture--checkerboard) transformations
that minimize the reprojection error in the image plane:
\[
\min_{\mX,\mZ}\sum_{mn}d^{2}\bigl(\bx_{mn},\hat{\bx}_{mn}(\mX,\mZ;\mA^{\prime}_{m},\bP_{n},\Kint)\bigr),
\]
where $d^{2}(\bx_{mn},\hat{\bx}_{mn})$ is the squared Euclidean distance between 
the measured projection $\bx_{mn}$ of the \mbox{$n$-th} checkerboard corner $\bP_n$ on the \mbox{$m$-th} camera
and the predicted corner $\hat{\bx}_{mn} = \mathbf{f}(\hat{\mB}^{\prime}_{m};\bP_{n},\Kint)$,
which is a function of the intrinsic camera parameters $\Kint$ and 
the extrinsic parameters 
$\hat{\mB}^{\prime}_{m}:=\mZ^{-1}\mA^{\prime}_{m}\mX$
predicted using the motion-capture data.
This non-linear least-squares problem is solved iteratively using the Gauss-Newton method. 
The initial value of $\mZ$ is given by 
$\mZ = \mA_1^{\prime} \mX \mB_1^{\prime-1}$, 
with $\mX$ provided by the above-mentioned linear algorithm.
We included a dataset for post-calibration in case another method is preferred.

\label{sec:ConvertGroundTruthPose}
The ground-truth pose gives the position and orientation of the event camera with respect to the world (i.e., the motion-capture system). 
Hence, it already incorporates the computed hand-eye transformation.
That is, while the motion-capture system outputs $\mT_{W H_j}$, we apply the hand-eye calibration $\mT_{H E}\equiv\mT_{H_j E_j}\;\forall j$ and directly report $\mT_{W E_j}=\mT_{W H_j} \mT_{H_j E_j}$ as ground-truth pose.

\subsection{Camera-IMU Calibration}\label{sec:calib-camera-imu}
The \texttt{calibration} dataset can be used to compute the Euclidean transformation between the camera and IMU reference frames. 
Running the publicly available software Kalibr~\citep{Furgale13iros} on the \texttt{calibration} dataset provides such a transformation, from the camera (i.e., the ``eye'' $E$) to the IMU, given by
\begin{equation*}
\mT_{\text{IMU},E} \approx
\begin{pmatrix}
0.9999 & -0.0122 & 0.0063 & 0.0067\\
0.0121 & 0.9998 & 0.0093 & 0.0007\\
-0.0064 & -0.0092 & 0.9999 & 0.0342\\
0 & 0 & 0 & 1
\end{pmatrix},
\end{equation*}
that is, the rotation matrix is approximately the identity (i.e., camera and IMU axes have the same orientation)
and the translation is dominantly along the optical axis $\approx$\SI{3.42}{\centi\meter}
(the IMU is a couple of cm behind the camera's optical center for the used lens).
Additionally, due to the IMU's built-in low-pass filter, 
the IMU measurements lag $\approx$\SI{2.4}{\milli\second} behind the images (and the events). 
This temporal shift is also reported by Kalibr.

\section{Known Issues}
\subsection{Clock Drift and Offset}
The clocks from motion-capture system and the DAVIS are not hardware-synchronized.
We observed clock drift of about \SI{2}{\milli\second/\minute}.
To counteract the clock drift, we reset the clocks before each dataset recording.
Since all datasets are rather short (in the order of \SI{1}{\minute}), the effect of drift is negligible. 
A small, dataset-dependent offset between the DAVIS and motion-capture timestamps is present since the timestamps were reset in software.

\balance
\bibliographystyle{SageH}
\bibliography{all}

\end{document}